\documentclass{article}

\usepackage{microtype}
\usepackage{graphicx}

\usepackage{booktabs} % for professional tables

\usepackage{hyperref}
\hypersetup{final}
\usepackage{subfig}

\usepackage{amsfonts}       % blackboard math symbols
\usepackage{nicefrac}       % compact symbols for 1/2, etc.
\usepackage{amsmath}
\usepackage[capitalize,nameinlink]{cleveref}

\usepackage[colorinlistoftodos]{todonotes}

\usepackage[accepted]{icml2020}

\usepackage{mathabx}

\usepackage{amssymb}

\usepackage{multicol}
\usepackage{multirow}
\usepackage{caption}

\def\schemeNameDFG{{\it DFG}}
\def\schemeNameResilinet{{\it ResiliNet}}
\def\failout{{\it failout}}

\def\health{health activity classification}
\def\cifar{CIFAR-10}

\def\low{{\em Hazardous}}
\def\med{{\em Poor}}
\def\high{{\em Normal}}

\newcommand{\subscriptscale}{0.8}

\begin{document}

\icmltitlerunning{ResiliNet: Failure-Resilient Inference in Distributed Neural Networks}

\twocolumn[
\icmltitle{ResiliNet: Failure-Resilient Inference in Distributed Neural Networks}




\begin{icmlauthorlist}
\icmlauthor{Ashkan Yousefpour}{fb}
\icmlauthor{Brian~Q.~Nguyen}{utd}
\icmlauthor{Siddartha~Devic}{utd}
\icmlauthor{Guanhua~Wang}{ucb}\\
\icmlauthor{Aboudy~Kreidieh}{ucb}
\icmlauthor{Hans Lobel}{puc}
\icmlauthor{Alexandre~M.~Bayen}{ucb}
\icmlauthor{Jason~P.~Jue}{utd}
\end{icmlauthorlist}

\icmlaffiliation{fb}{Facebook AI, California, USA. Part of this work was done when the author was at UC Berkeley and UT Dallas.}
\icmlaffiliation{utd}{UT Dallas, Texas, USA}
\icmlaffiliation{ucb}{UC Berkeley, California, USA}
\icmlaffiliation{puc}{PUC Chile, Santiago, Chile}

\icmlcorrespondingauthor{Ashkan Yousefpour}{yousefpour@fb.com}

\icmlkeywords{Deep Learning, Distributed Inference, Distributed Neural NetworksFailure-Resiliency}

\vskip 0.3in
]



\printAffiliationsAndNotice{} % otherwise use the standard text.

\begin{abstract}
Federated Learning aims to train distributed deep models without sharing the raw data with the centralized server. Similarly, in distributed inference of neural networks, by partitioning the network and distributing it across several physical nodes, activations and gradients are exchanged between physical nodes, rather than raw data. Nevertheless, when a neural network is partitioned and distributed among physical nodes, failure of physical nodes causes the failure of the neural units that are placed on those nodes, which results in a significant performance drop. Current approaches focus on resiliency of training in distributed neural networks. However, resiliency of inference in distributed neural networks is less explored. We introduce \schemeNameResilinet{}, a scheme for making inference in distributed neural networks resilient to physical node failures. \schemeNameResilinet{} combines two concepts to provide resiliency: {\em skip hyperconnection}, a concept for skipping nodes in distributed neural networks similar to skip connection in resnets, and a novel technique called \failout{}, which is introduced in this paper. Failout simulates physical node failure conditions during training using dropout, and is specifically designed to improve the resiliency of distributed neural networks. The results of the experiments and ablation studies using three datasets confirm the ability of \schemeNameResilinet{} to provide inference resiliency for distributed neural networks.
\end{abstract}

\section{Introduction}
% MOTIVATION (standard stuff, just like the other paper, but more focused on ML)

% PROBLEM (distributed learning, moving to edge-fog-cloud)

% CURRENT SOLUTIONS

Deep neural networks (DNNs) have boosted the state-of-the-art performance in various domains, such as image classification, segmentation, natural language processing, and speech recognition \cite{krizhevsky2012imagenet,hinton2012deep,lecun2015deep, nlp}. In certain DNN-empowered IoT applications, such as image-based defect detection or recognition of parts during product assembly, or anomaly behavior detection in a crowd, the {\em inference} task is intended to run for a {\em prolonged period of time}, that is, there is a continuous stream of samples in inference. In these applications, a recent trend \cite{harvard, hotedge-distributed} has been to partition and {\em distribute} the computation graph of a previously-trained neural network over physical nodes along an edge-to-cloud path (e.g. on {\em edge} and {\em fog} servers) so that the forward-propagation occurs {\em in-network} while the data traverses toward the cloud. This inference in distributed DNN architecture is motivated by two observations: Firstly, deploying DNNs directly onto IoT devices for huge multiply-add operations is often infeasible, as many IoT devices are low-powered and resource-constrained \cite{zhou2019lightweight}. Secondly, placing the DNNs in the cloud may not be reasonable for such prolonged inference tasks, as the raw data, which is often large, has to be continuously transmitted from IoT devices to the DNN model in the cloud, which results in the high consumption of network resources and possible privacy concerns~\cite{ionn, harvard}. 

A natural question that arises within this setting is whether the inference task of a distributed DNN is resilient to the failure of individual physical nodes, since distributing a DNN over several machines induces a higher risk of failures. The failure model in this paper is {\em crash-only non-Byzantine}; physical nodes could fail due to power outages, cable cuts, natural disasters, or hardware/software failures. Providing failure-resiliency for such inference tasks is vital, as physical node failures are more probable during a long-running inference task (e.g., in a continuous stream of samples). Failure of a physical node causes the failure of the DNN units that are placed on the node, and is especially troublesome for IoT applications that cannot tolerate poor performance while the physical node is being recovered.
% time, e.g. automated vehicle platooning, live shopping recommendation,  applications~\cite{franke1995truck, bergenhem2012overview}. 
The following question is the topic of our study. 

{\em How can we make distributed DNN inference resilient to physical node failures?} 

Several frameworks have been developed for distributed training of neural networks \cite{tensorFlow, Pytorch, microsoft}. On the other hand, inference in distributed DNNs has emerged as an approach for DNN-empowered IoT applications. Providing failure-resiliency during inference for such IoT applications is crucial. Authors in \cite{previous} introduce the concept of {\em skip hyperconnections} in distributed DNNs that provides some failure-resiliency for inference in distributed DNNs. Skip hyperconnections skip one or more {\em physical} nodes in the vertical hierarchy of a distributed DNN. These forward connections between non-subsequent physical nodes help in making distributed DNNs more resilient to physical failures, as they provide alternative pathways for information when a physical node has failed. Although superficially they might seem similar to skip connections in residual networks \cite{ms-residual}, skip hyperconnections serve a completely different purpose. While the former aim at solving the vanishing gradient problem during training, the latter are based on the underlying insight that during inference, if at least a part of the incoming information for a physical node is present (via skip hyperconnections), given their generalization power, the neural network may be able to provide a reasonable output, thus providing failure-resiliency.

% LIMITATIONS
A key observation in the aforementioned work is that the weights learned during training using skip hyperconnections are not \textit{aware} that there might be physical node failures. In other words, the information about failure of physical nodes is not used during training to make the learned weights aware of such failures. As such, skip hyperconnections by themselves do not make the learned weights more resilient to physical failures, as they are just a way to diminish the effects of losing the information flow at inference time.

{\bf Key Contributions}: Motivated by this limitation, (1) we introduce \schemeNameResilinet{}, which utilizes a new regularization scheme we call \textit{failout}, in addition to skip hyperconnections, for making inference in distributed DNNs resilient to physical node failures. Failout is a regularization technique that during training ``{\em fails}'' (i.e. shuts down) the physical nodes of the distributed DNN, each hosting several neural network layers, thus simulating inference failure conditions. Failout effectively embeds a resiliency mechanism into the learned weights of the DNN, as it forces the use of skip hyperconnections during failure. The training procedure using failout could be applied offline, and would not necessarily be done during runtime (hence, shutting down physical nodes would be doing so in simulation). Although in \schemeNameDFG{} framework \cite{previous} skip hyperconnections are always active both during training and inference, in \schemeNameResilinet{} skip hyperconnections are active during training and during inference only when the physical node that they bypass fails (for bandwidth savings).  

(2) Through experiments using three datasets we show that \schemeNameResilinet{} minimizes the degradation impact of physical node failures during inference, under several failure conditions and network structures. Finally, (3) through extensive ablation studies, we explore the rate of failout, the weight of hyperconnections, and the sensitivity of skip hyperconnections in distributed DNNs. \schemeNameResilinet{}'s major novelty is in providing failure-resiliency through special training procedures, rather than traditional ``system-based'' approaches of redundancy, such as physical node replication or backup.

\section{Resiliency-based Regularization for DNNs} \label{definitions}
%In this section we introduce the building blocks of the \schemeNameResilinet{} architecture, namely distributed neural networks, skip hyperconnections, and failout regularization.
%  We use boldface lower-case letters for vectors (e.g. $\mathbf{x}$, $\mathbf{w}$) and boldface upper-case letters for matrices (e.g. $\mathbf{W}$). $\mathbf{0}$ and $\mathbf{1}$ denote vectors of zeros and ones, respectively. 

\subsection{Distributed neural networks}
A distributed DNN is a DNN that is split according to a partition map and distributed over a set of physical nodes (a form of model parallelism). This concept is sometimes referred to as {\em split learning}, where only activations and gradients are transferred in the distributed DNN, which can result in improvements in privacy. This article studies the resiliency of {\em previously-partitioned} distributed DNN models during inference. We do not study the problem of optimal partitioning of a DNN; the optimal DNN partitioning depends on factors such as available network bandwidth, type of DNN layers, and the neural network topology \cite{ChuangHu, neurosurgeon, zhou2019distributing}. We do not consider doing any neural architecture search in this article. Nevertheless, in our experiments, we consider different partitions of the DNNs.

Since a distributed DNN resides on different physical nodes, during inference, the vector of output values from one physical node must be transferred (e.g. through a TCP socket) to another physical node. The transfer link (pipe) between two physical nodes is called a {\em hyperconnection} \cite{previous}. Hyperconnections transfer information (e.g. feature maps) as in traditional connections between neural network layers, but through a physical communication network. Unlike a typical neural network connection that connects two units and transfers a scalar, a hyperconnection connects two physical nodes and transfers a vector of scalars. Hyperconnections are one of two types: simple or skip. A simple hyperconnection connects a physical node to the physical node that has the next DNN layer. Skip hyperconnections are explained next.

\subsection{Skip Hyperconnections}
The concept of skip hyperconnections is similar to that of skip connections in residual networks (ResNets) \cite{ms-residual}. A skip hyperconnection \cite{previous} is a hyperconnection that skips one or more physical nodes in a distributed neural network, forwarding the information to a physical node that is further away in the distributed neural network structure. During training, the DNN learns to use the skip hyperconnections to allow a downstream physical node (closer to cloud) receive information from more than one upstream physical node. Consequently, during inference, if a physical node fails, information from the prior working nodes are still capable of propagating forward to downstream working physical nodes via these skip hyperconnections, providing some failure-resiliency \cite{previous}.

\schemeNameResilinet{} also uses skip hyperconnections, but in a slightly different manner from the \schemeNameDFG{} framework. When there is no failure during inference, or no failout during training (failout, to be discussed), the skip hyperconnections are not active. When failure occurs during inference (failures can be detected by simple heartbeat mechanisms), or failout during training, skip hyperconnections become active, to route the {\em blocked} information flow. This setup in \schemeNameResilinet{} significantly saves bandwidth, compared to \schemeNameDFG{}, which requires skip hyperconnections to be always active. The advantage here is in routing information during failure, that is otherwise not possible. Also, the bandwidth for the routed information over the failed node is the same as when there is no failure (skip hyperconnection only finds a detour). In the experiment we also show through experiments that if skip hyperconnections are always active, the performance only increases negligibly. 

\subsection{Failout Regularization}
In the \schemeNameDFG{} framework \cite{previous}, the information regarding failure of the physical nodes is not used during training to make the learned weights more aware of such failures. Although skip hyperconnections increase the failure-resiliency of distributed DNNs, they do not make the learned weights more prepared for such failures. This is because all neural network components are present during training, as opposed to inference time where some physical nodes may fail. In order to account for the learned weights being more adapted to specific failure scenarios, we introduce failout regularization, which simulates inference-time physical node failure conditions during training. 

During training, failout ``{\em fails}'' (i.e., shuts down) a physical node, to make the learned weights more adaptive to such failures and the distributed neural network more failure-resilient. By ``failing'' a physical node, we mean temporarily removing the neural network components that reside on the physical node, along with all their incoming and outgoing connections. Failout's training procedure could be done offline, and would not necessarily be employed during runtime. Therefore failing physical nodes would be temporarily removing their neural network components in simulation.

When the neural components of a given physical node shut down using failout, the neural layers of the downstream physical node that are connected to the failing physical node will not receive information from the failing physical node, forcing their weights to take into account this situation and utilize the received information from the skip hyperconnection. In other words, failout forces the information passage through the skip hyperconnections during training, hence adapting the weights of the neural network to account for these failure scenarios during inference. 

% can-remove-for-space
%Note that if a physical node does not have a skip hyperconnection bypassing it, applying failout on it during training will not improve the failure-resiliency, as there is no alternative information path if the physical node fails. Thus, it is not required to apply failout on physical nodes that do not have skip hyperconnections bypassing them. 

Formally, consider a neural network which is distributed over $V$ different nodes $v_i$, $i \in [1,V]$, where for each $v_i$, we define its failure rate (probability of failure) $f_i \in [0,1]$. Following this, we define a binary mask $b$ with $V$ components, where its $i$-th element $b_i$ follows a Bernoulli distribution, with a mean equal to $1-f_i$, that is $b_i \sim Ber(1-f_i)$. During training, for each batch of examples, a new mask $b$ is sampled, and if $b_i=0$, the neural components of physical node $v_i$ are dropped from computation ($v_i$'s output is set to zero in simulated off-line training), thus simulating a real failure scenario. Formally, if $Y_i$ denotes the output of node $v_i$, then $Y_i=b_iH_i(X_i)$, where $H_i(.)$ is a non-linear transform on $X_i$, the input of physical node $v_i$. 

Consider a vertically distributed DNN where the nodes are numbered in sequence $1,2,\ldots,V$ from upstream to downstream (cloud). In this setting, for \schemeNameResilinet{} we can derive an equation for $X_{i+1}$, the input to node $v_i$, as
\begin{equation}
    X_{i+1}=Y_i\odot Y_{i-1}, \label{main-eq}
\end{equation}
where the operator $\odot$ is defined as: in $X_{i+1}=Y_i\odot Y_{i-1}$, when node $v_i$ is alive $X_{i+1}=Y_i$, and when node $v_i$ fails, $X_{i+1}=Y_{i-1}$. In this definition, we can see that the skip hyperconnection is only active when there is a failure, which corresponds to \schemeNameResilinet{}. In the experiments, we also consider a case where skip hyperconnections are always active (we call it \schemeNameResilinet+), for which \cref{main-eq} is modified to 
\begin{equation}
    X_{i+1}=Y_i\oplus Y_{i-1}.
\end{equation}

Although, superficially, failout seems similar to dropout \cite{normal-dropout}, failout removes a whole segment of neural components, including neurons and weights, for a different purpose of failure-resiliency of distributed DNNs (and not only regularizing the neural network). Another distinction between failout and dropout is in their behavior during inference. In dropout, at inference, the weights of connections are multiplied by the probability of survival of their source neuron to account for model averaging from exponentially many thinned models. Furthermore, DNN units are not dropped during inference in dropout, making the model averaging a necessity. In contrast, failout does not multiply weights of hyperconnections by the probability of survival, since, during inference, physical nodes may fail, though not necessarily at the same rate as during training. Said differently, failout does not use the model ensemble analogy as used in standard dropout, hence does not need the mixed results of the model ensembles. To verify our hypothesis, we conducted experiments using different datasets in a setting where the weights of hyperconnections are multiplied by the probability of survival of the physical nodes, and we observed a sheer reduction in performance.

\section{Experiments} \label{new_experiment}
We compare \schemeNameResilinet{}'s performance with that of \schemeNameDFG{} \cite{previous} and {\em vanilla} (distributed DNN with no skip hyperconnections and no failout). 

\subsection{Scenarios and Datasets}
%We evaluate the resiliency of our approach in two relevant distributed DNN scenarios: vertically distributed MLP and vertically distributed CNN. We first describe each of the scenarios and the datasets used for each scenario. 

\begin{figure}[t]
    \centering
\begin{minipage}[b]{0.42\linewidth}
    \centering
    \includegraphics[width=\linewidth]{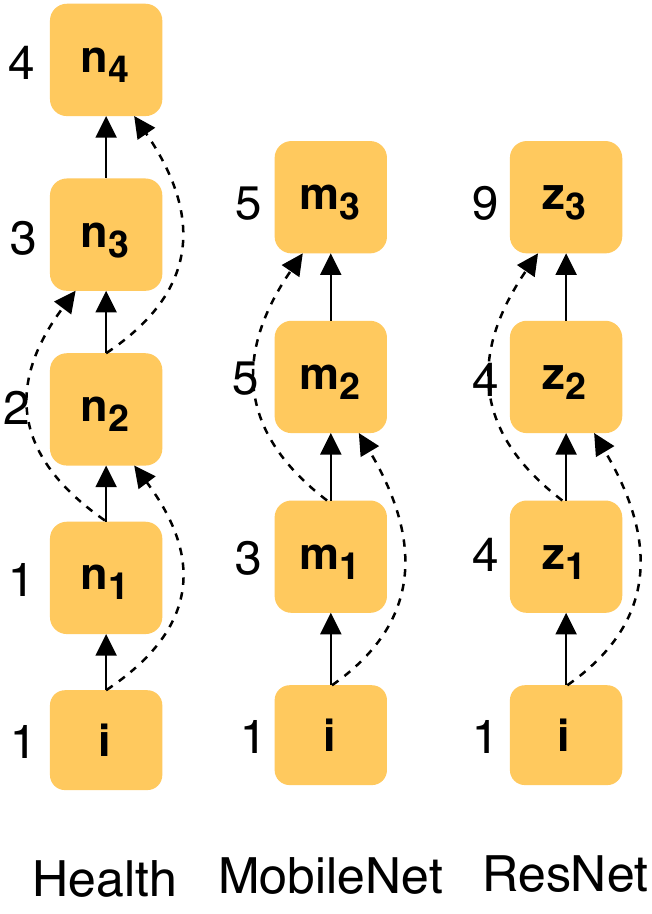}
    \caption{Distributed neural network setup and number of layers on each node.}
    \label{fig:experiment-setting}
\end{minipage}
\begin{minipage}[b]{0.56\linewidth}
    \centering
    \includegraphics[width=\linewidth]{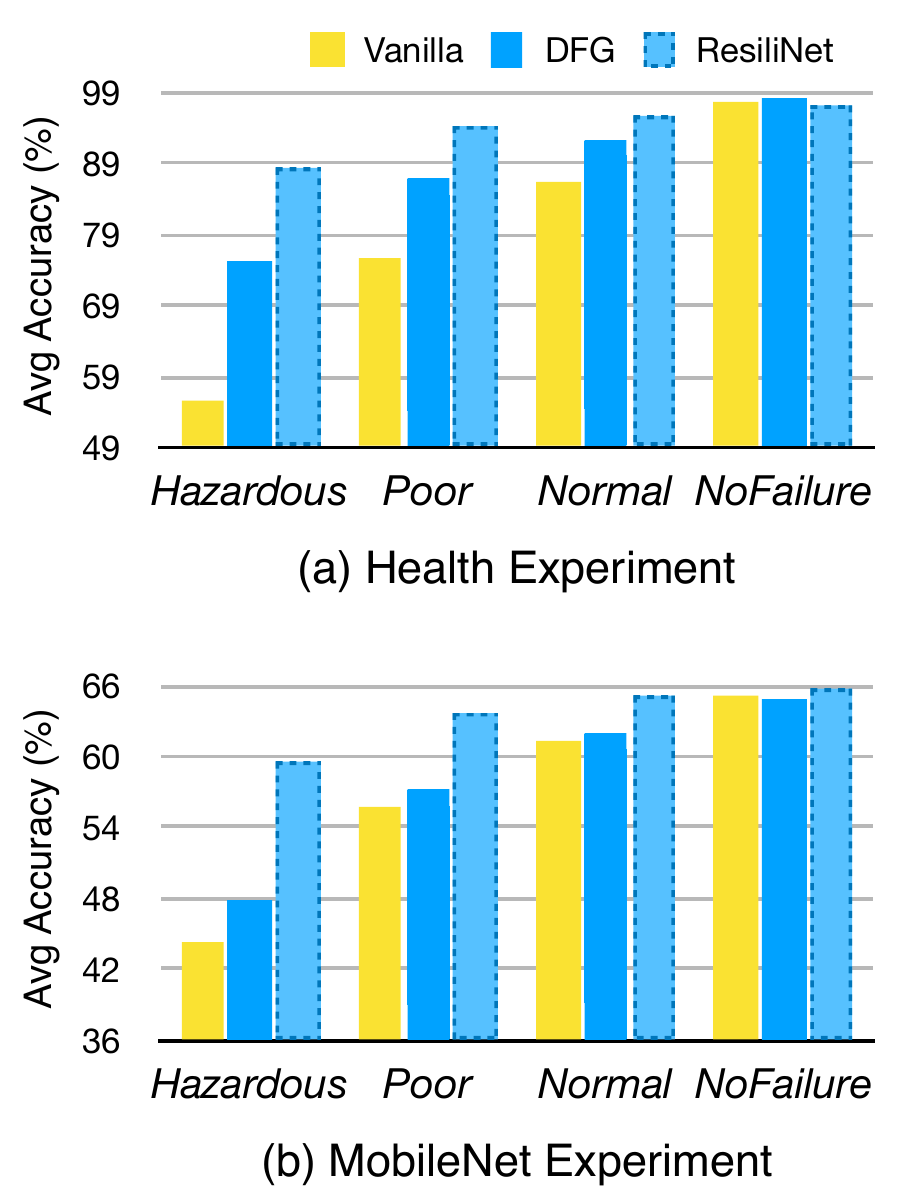}
    \caption{Average performance}
    \label{fig:main-results}
\end{minipage}
\end{figure}

{\bf Vertically distributed MLP}:
This is the simplest scenario for a distributed DNN in which the MLP is split vertically across physical nodes shown in the left of \cref{fig:experiment-setting}.  

For this scenario, we use the UCI {\em \health{}} dataset (``Health'' for short), described in \cite{mhealth}. This dataset is an example of an IoT application for medical purposes where the inference task will run over a long period of time. The dataset is comprised of readings from various sensors placed at the chest, left ankle, and right arm of 10 patients. There are a total of 23 features, each corresponding to a type of data collected from sensors. For this experiment, we split a DNN that consists of ten hidden layers of width 250, over 4 physical nodes as follows. The physical node $n_1$ hosts one hidden layer, $n_2$ two, $n_3$ three, and $n_4$ four (also summarized in \cref{tab:surv-configs}). 
The dataset is labeled with the 12 activities a patient is performing at a given time, and the task is to classify the type of activity. We remove the activities that do not belong to one of the classes. After reprocessing, the dataset has 343,185 data points and is roughly uniformly distributed across each class. Hence, we use a standard cross-entropy loss function for the classification. For evaluation, we separate data into train, validation, and test with an 80/10/10 split.

{\bf Vertically distributed CNN}: The two architectures in the right in \cref{fig:experiment-setting} present the neural network structure proposed for these scenarios and how the CNNs are split. For these scenarios, we use two datasets, ImageNet Large Scale Visual Recognition Challenge ({\em ILSVRC}) and CIFAR-10. We utilize the ILSVRC dataset for measuring the performance of \schemeNameResilinet{} in distributed CNNs. However, for ablation studies for distributed CNNs, we use the CIFAR-10 dataset, since we run several iterations of experiments with different hyperparameters. We also employ data augmentation to improve model generalization. 

% The {\em CIFAR-10} dataset consists of 60,000 images that correspond to 10 objects \cite{Krizhevsky09}. The dataset is split into 50,000 examples for training and 10,000 for testing. In both datasets, the distribution of data points is uniform, hence we used a cross-entropy loss for training. 

For CIFAR-10 and ImageNet datasets, we use the MobileNetV1 CNN architecture ~\cite{howard2017mobilenets} and split it across 3 physical nodes. We chose version 1 of MobileNet (MobileNetV1), as it does not have any of the skip connections that are present in MobileNetV2. Moreover, since non-residual models cannot effectively deal with layer failure \cite{residual2}, we also consider neural networks with residual connections and experiment with ResNet-18 \cite{ms-residual}. The ResNet-18 architecture has 18 layers, and we partition these stacked layers across the three physical nodes: $z_1$ contains four layers, $z_2$ four, and $z_3$ 9 plus the remaining layers. The MobileNetV1 architecture has 13 ``stacked layers'', each with the following six layers: depth-wise convolution, batch normalization, ReLU, convolution, batch normalization, and ReLU. We partition these 13 stacked layers across the three physical nodes.
%: $m_1$ contains three stacked layers, $m_2$ five, and $m_3$ five plus the remaining layers (average pooling, dropout, conv2D, and softmax). 

\subsection{Experiment Settings}
%We implemented our experiments using TensorFlow and Keras on Amazon Web Service EC2 \texttt{p3.8xlarge} instance, which has 4 GPUs, 32 vCPUs, and 244 GiB RAM. 
%In the neural network architectures, we only include the skip hyperconnections that bypass one physical node, since, in the edge-fog-cloud networks, the number of physical nodes over which the DNN is distributed is small. 
%Moreover, we experimented with models having skip hyperconnections that skip more than one physical node, but did not observe any performance gain.

We implemented our experiments using TensorFlow and Keras on Amazon Web Service EC2 instances. Batch sizes of 1024, 128, and 1024 are used for the Health, CIFAR-10, and ImageNet experiments, respectively. The learning rate of 0.001 is used for the \health{} and \cifar{} experiments. Learning rate decay with an initial rate of 0.01 is used for the ImageNet experiment. The image size of $160\times 160\times 3$ pixels is used for the ImageNet experiment. The rate of failout for \schemeNameResilinet{} is set to $10\%$ (other rates of failout are explored later in ablation studies).

{\bf Failure probabilities}: To empirically evaluate different schemes, we use three different {\em failure settings} outlined in \cref{tab:surv-configs}. A failure setting is a tuple, where each element $i$ is the probability that the physical node $v_i$ fails during inference. For example, the setting \high{} could represent more reasonable network condition, where the probability of failure is low, while the settings \med{} and \low{} represent failure settings (only for experiments) when the failures are very frequent in the physical network. It is worth noting that the specific values of failure probabilities do not change the overall trend in the results and are only chosen so we have some benchmark for three different failure conditions. 

To obtain values for the failure probabilities, we have the following observations: 1) the top physical node ($n_4$, $m_3$, $z_3$ in \cref{fig:experiment-setting}) is the cloud, and hence is always available; 2) the nodes closer to the cloud are more available than the ones far from the cloud; 3) the physical nodes closer to the cloud and data centers (e.g., backbone nodes) have relatively high availability of around 98\% \cite{meza2018large}: such nodes have a mean time between failures (MTBF) of 3521 hours and a mean time to repair (MTTR) of 71 hours. Thus, the availability of those nodes is around 98\%, while presumably the physical nodes closer to end-user are expected to have less availability, of around 92\%-98\% (in failure setting \high{}). Once we obtain values for failure setting \high{}, we simply increase them for settings \med{} and \low{}. 
% Since we are studying the resiliency of distributed DNNs (and not their input), we did not consider the failure of input.

% can-remove-for-space
%When performing operations on the output of hyperconnections, it may be necessary to match the dimensions of the hyperconnections. For distributed MLPs (multi-layer perceptrons), to change the dimension of the hyperconnections, we can do zero-padding or use a fully-connected layer. For distributed CNNs (convolutional neural networks), we can do zero-padding or $1\times 1$ convolutions to change the dimensions~\cite{ms-residual, further-ablation-studies}.

\subsection{Performance Evaluation}

\begin{table}[t] 
    \centering
\scriptsize
\vspace{-5pt}
\centering
\setlength{\tabcolsep}{2pt}
\begin{tabular}{@{}clc|cccc@{}}
\toprule
& \multicolumn{2}{c}{Failing} & \multicolumn{4}{c}{Top-1 Accuracy (\%)}   \\
& Nodes & Prob. ($\%$) & ResiliNet+ & ResiliNet & DFG & Vanilla \\ \midrule
% \midrule
% \multicolumn{7}{c}{Health Experiment}\\
\midrule
% & \multicolumn{2}{c}{~} & \multicolumn{3}{c}{Accuracy (\%)}   \\
\parbox[t]{2mm}{\multirow{9}{*}{\rotatebox[origin=c]{90}{{\scriptsize Health}}}}
& None & 87.43 & 97.85 & 97.77 & \textbf{97.90} & 97.85 \\
& $n_1$& 7.01 & \textbf{97.35} & 93.26 & 64.42      & 7.95 \\
& $n_2$  & 3.64& 94.32 & \textbf{95.59} & 22.49     & 7.99 \\
& $n_3$ & 0.88 & \textbf{97.74} & 97.12 & 92.48     & 8.10 \\
& $n_1, n_2$ & 0.32 & 8.02 & 8.12 & 8.2  & 7.93 \\
& $n_1, n_3$ & 0.08 & \textbf{97.33} & 91.12 & 60.13     & 7.98 \\
& $n_2, n_3$ & 0.04& 7.99 & 7.86 & 7.98     & 7.97 \\
 & $n_1, n_2, n_3$ & 0.003 & 7.98 &  8.11 & 7.89     & 7.91 \\[0.3em]
%   \midrule
 & Average & ~ & \textbf{97.36} & 97.02 & 92.21 & 86.57  \\
\midrule
% & \multicolumn{2}{c}{~} & \multicolumn{3}{c}{F-measure (\%)}   \\ 
% \parbox[t]{2mm}{\multirow{9}{*}{\rotatebox[origin=c]{90}{{\scriptsize Camera}}}}
% & None & 88.06 & \textbf{99.09}       &  98.53     &  98.86  \\
% & $z_2$ & 1.80 & \textbf{98.10} & 97.56    & 98.04 \\
% %& $z_3$ & 1.80 & 94.39 & \textbf{96.83}      & 98.13 \\
% & $z_5$ & 0.89 & \textbf{97.74} & 96.86     & 96.99   \\
% & $z_6$ & 0.89 & 96.65 & \textbf{97.64}      & 89.43 \\
% & $z_7$ & 0.89 & \textbf{95.39} & 91.30  & 59.34 \\
% & $z_8$ & 0.89 & \textbf{98.46} & 97.88      & 60.24 \\
% & $z_2, z_7$ & 0.02 & \textbf{91.68} & 85.44  & 61.46 \\
% & $z_1, z_7$ & 0.02 & \textbf{91.32} & 79.51  & 59.43 \\
% & $z_1, z_4, z_6$  & 0.001  & \textbf{93.04}  & 85.77 & 60.56  \\
% & All & 0.001 & 59.18 & 61.62  & 60.89 \\
% \midrule
% \multicolumn{7}{c}{CIFAR Experiment}\\
%  & \multicolumn{2}{c}{~} & \multicolumn{3}{c}{Accuracy (\%)}   \\
\parbox[t]{2mm}{\multirow{5}{*}{\rotatebox[origin=c]{90}{{\scriptsize MobileNet}}}}
& None & 94.08 & \textbf{88.11} & 87.75 & 87.54 & 86.64 \\
& $m_1$ & 3.92 & \textbf{78.98} & 75.55 & 69.42 & 10.27 \\
& $m_2$ & 1.92 & \textbf{75.65} & 59.18 & 62.76 & 9.85 \\
& $m_1, m_2$  & 0.08 & 9.71 & 10.11 & 10.02 & 10.07 \\[0.3em]
% \midrule
%  & \multicolumn{2}{c}{~} & \multicolumn{3}{c}{Accuracy (\%)}   \\
% \parbox[t]{2mm}{\multirow{3}{*}{\rotatebox[origin=c]{90}{{\scriptsize ResNet}}}}
% & None & 94.08 & 90.06 & \textbf{92.41} & 92.26 \\
% & $z_1$ & 3.92 & \textbf{56.24} & 24.57 & 9.16 \\
% & $z_2$ & 1.92 & \textbf{88.01} & 70.31 & 9.66 \\
% & $z_1, z_2$  & 0.08 & 9.91 & 9.58 & 9.45 \\
 & Average &  & \textbf{87.45} & 86.66 & 86.29 & 82.1  \\
 \bottomrule
\end{tabular}

\caption{Individual physical node failures}
\label{tab:node-failure}
\end{table}

\Cref{tab:node-failure} shows the performance of different schemes for certain physical node failures. The first two columns show the failing nodes, along with the probability of occurrence of those node failures under \high{} failure setting. Recall that Vanilla is a distributed DNN that does not have skip hyperconnections and does not use failout. We assume that, when there is no information available to do the classification task due to failures, we do random guessing. \schemeNameResilinet{}+ is a scheme based on \schemeNameResilinet{} where skip hyperconnections are always active, during inference (or validation) and training. (In this table, for MobileNet experiment \cifar{} dataset is used).

\begin{figure*}[t]
    \centering
    
    \begin{minipage}[b]{0.57\linewidth}
    \captionsetup{type=table}
    \centering
    % \vspace{-1in}
\setlength{\tabcolsep}{10pt}
\tiny
\caption{Experiment settings}
% \vspace{-0.1in}
\begin{tabular}{@{}lccc@{}}
\toprule
Experiment    & Dist. MLP & Dist. MobileNet & Dist. ResNet-18\\ \midrule
Dataset       & UCI Health & ImageNet, CIFAR-10 & CIFAR-10\\ \midrule
Nodes Order & $[n_4, n_3, n_2, n_1]$ & $[m_3, m_2, m_1]$ & $[z_3, z_2, z_1]$\\ \midrule
%\multicolumn{4}{c}{{\bf Number of DNN layers hosted on each physical node}} \\ \midrule
%~~~& $[4, 3, 2, 1]$ & $[5, 5, 3]$ & $[9, 4, 4]$\\\midrule

% \multicolumn{4}{c}{{\bf Probability of failure}} \\ \midrule
 Failure Setting &  \\
~~~{\em Normal}& $[0\%, 1\%, 4\%, 8\%]$ & $[0\%, 2\%, 4\%]$ & $[0\%, 2\%, 4\%]$\\
~~~{\em Poor}           & $[0\%, 5\%, 9\%, 13\%]$  & $[0\%, 5\%, 10\%]$  & $[0\%, 5\%, 10\%]$\\
~~~{\em Hazardous}              &[$0\%, 15\%, 20\%, 22\%]$ & $[0\%, 15\%, 20\%]$ & $[0\%, 15\%, 20\%]$\\
\bottomrule
\end{tabular}
\label{tab:surv-configs}
    \end{minipage}
    \begin{minipage}{.0\linewidth}
    ~
    \end{minipage}
    \begin{minipage}[b]{0.40\linewidth}
    \centering
\includegraphics[width=1.0\linewidth]{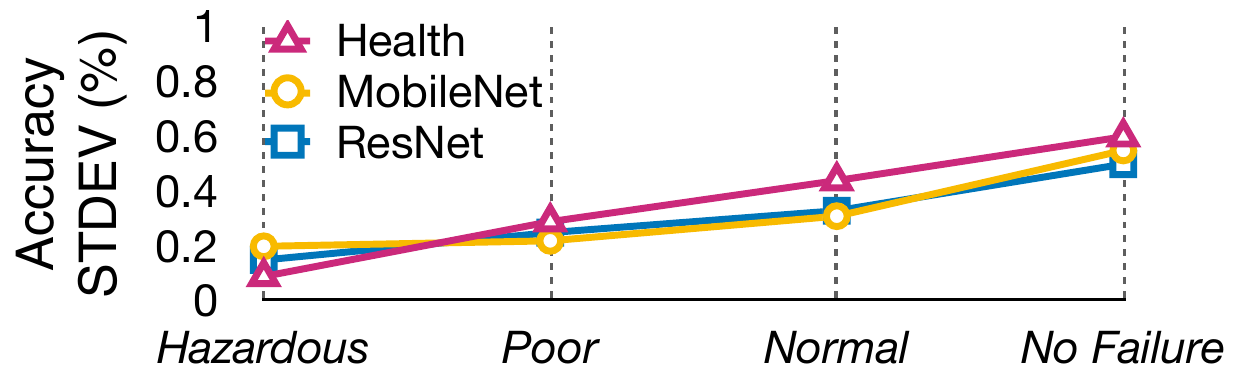}
    \caption{Impact of hypercon. weight in \schemeNameResilinet{}}
     \vspace{-0.4in}
    \label{fig:experiment-weight}
    \end{minipage}

\end{figure*}

{\textbf{(a) Health}}: In the \health{} experiment, we see that the failure of even a single physical node compromises the performance of Vanilla due to random guessing, resulting top-1 accuracy of around 8\%. On the other hand, \schemeNameDFG{}, \schemeNameResilinet{}, and \schemeNameResilinet{}+ subvert Vanilla's inability to pass data over failed physical nodes, thereby achieving significantly greater performance. The results also show that, in this experiment, \schemeNameResilinet{} and \schemeNameResilinet{}+ perform better than \schemeNameDFG{} in all of the cases, except for when there is no failure. In certain physical nodes failures, such as when $n_1$, $n_2$, or $\{n_1, n_3\}$ fail, \schemeNameResilinet{} and \schemeNameResilinet{}+ greatly surpass the accuracy of the both \schemeNameDFG{} and Vanilla, providing a high level of failure-resiliency. When physical node failures $\{n_1, n_2\}$ and $\{n_2, n_3\}$ occur, all schemes do not provide high accuracy, due to inaccessibility of the path for information flow.

{\bf (b) MobileNet on \cifar}: In the MobileNet experiment with \cifar{} dataset, Vanilla is outperformed by other three schemes when there is any combination of failures. \schemeNameResilinet{} and \schemeNameResilinet{}+ both offer a great performance when $m_1$ fails; nevertheless, \schemeNameDFG{} performs marginally better than \schemeNameResilinet{} when $m_2$ fails. \schemeNameResilinet{}+ consistently has the highest accuracy in this experiment. 

We can see that \schemeNameResilinet{} overall maintains a higher accuracy than \schemeNameDFG{} and vanilla. We can also see that \schemeNameResilinet+ outperforms all of the schemes. However, this benefit comes at a cost of having the skip hyperconnections always active, which results in higher bandwidth usage. In the rest of the experiments, we choose \schemeNameResilinet{} among the two \schemeNameResilinet{}{\em s}. This is a pessimistic choice and it is justified by the bandwidth savings.

Previously, we discussed and showed how the {\em accuracy} is affected when particular physical nodes fail. Nevertheless, some of the physical node failures are not as probable as others (e.g. multiple physical nodes failure vs. single physical node failure), and hence it is interesting to see the {\em average accuracy} in different node failure settings. \cref{fig:main-results} shows the average top-1 accuracy of the three methods under different failure settings, with 10 iterations for the \health{} experiment, and 2 iterations for the MobileNet experiment on ImageNet (confidence intervals are very small and negligible and are omitted). {\bf\em Key result 1}: as expected, in both experiments, \schemeNameResilinet{} seems to outperform \schemeNameDFG{} and Vanilla. The high performance of \schemeNameResilinet{} is more evident in severe node failure conditions.

\subsection{Ablation Studies} \label{ablation}

Now that the validity of failout has been empirically shown to provide an increase in failure-resiliency of distributed neural networks, we now investigate the importance of individual skip hyperconnections, their weights, as well as the optimal rate of failout. To do so, we raise four important questions in what follows and empirically provide answers to these questions. We use the CIFAR-10 dataset for ablation studies of the distributed CNN, and use Health for ablation studies of distributed MLPs. 

{\bf 1. \em What is the best choice of weights for the hyperconnections?} Hyperconnections can have weights, similar to the weights of the connections in neural networks. We begin by assessing the choice of weights of the hyperconnections. Although by default, the weight of hyperconnections in \schemeNameResilinet{} is $\mathbf{1}$, we pondered if setting the weights relative to the {\em reliability} of their source physical nodes could improve the accuracy. Reliability of a physical node $v_i$ is $r_i=(1-p_i)$, where $p_i$ is the probability of failure of node $v_i$. We proposed two heuristics, called ``Relative Reliability'' and ``Reliability,'' that are described as follows: \\
Consider two physical nodes $v_1$ and $v_2$ feeding data through hyperconnections to physical node $v_3$. If physical node $v_1$ is less reliable than physical node $v_2$ ($r_1<r_2$), setting $v_1$'s hyperconnections weight with a smaller value than that of $v_2$ may improve the performance. Thus, for the hyperconnection weight connecting node $v_i$ to node $v_j$, in Reliability heuristic, we set $\mathbf{\overline{\overline{w}}}_{ij}=r_i$, where $\mathbf{\overline{\overline{w}}}_{ij}$ denotes the weight of hyperconnection from physical node $v_i$ to node $v_j$. Comparably, in Relative Reliability heuristic, we set $\mathbf{\overline{\overline{w}}}_{ij}=\frac{r_i}{\sum_{k\in H_j}{r_{k}}}$, where $H_j$ is the set of incoming hyperconnection indices to the physical node $v_j$.

We experiment with the following four hyperconnection weight schemes in \schemeNameResilinet{} for 10 runs: (1) weight of $\mathbf{1}$, (2) Reliability heuristic, (3) Relative Reliability heuristic, and (4) uniform random weight between 0 and 1. {\bf\em Key result 2}: surprisingly, all of the four hyperconnection weight schemes resulted in a similar performance. Since all of the values for average accuracy are similar in these experiments, we report in \cref{fig:experiment-weight} the standard deviation among these weight schemes in \schemeNameResilinet{}. 

We see that the standard deviation among the weight schemes is negligible, constantly below 1\%. This suggests that there may not be a significant difference in accuracy when using any of the {\em reasonable} weighting scheme (e.g. heuristic of $\mathbf{1}$). {\bf\em Key observation 1}: we also experimented with a scheme in which the hyperconnection weight is uniformly and randomly distributed between 0 and 10, and observed that the accuracy dropped significantly for the distributed MLPs. {\bf\em Key observation 2}: surprisingly, the accuracy of distributed CNNs stays in the same range as in other schemes, when hyperconnection weight is a uniform random number between 0 and 10. We hypothesize that, for distributed MLPs, a {\em reasonable} hyperconnection weight scheme is a scheme that assigns the weights of hyperconnections between 0 and 1. Nevertheless, further investigation may be required in different distributed DNN architectures to assess the full effectiveness of hyperconnection weights.

\begin{table*}[t]
\centering
\scriptsize
\setlength{\tabcolsep}{4pt}
    % \vspace{-3.2in}
    \caption{Impact of failout rate in \schemeNameResilinet{}. Numbers represent average top-1 accuracy in \%.} 
    % \vspace{-5pt}
    \begin{tabular}{@{}l|ccc|ccc|ccc|ccc|ccc@{}}
    \toprule
    Failout Rate      &
    \multicolumn{3}{c|}{``Failure''} & \multicolumn{3}{c|}{5\%} & \multicolumn{3}{c|}{10\%} & \multicolumn{3}{c|}{30\%} & \multicolumn{3}{c}{50\%}  \\ \midrule
    Experiment & H  & M & R & H & M & R & H & M & R & H & M & R & H & M & R\\ \midrule
    Failure Setting & ~ & ~ & ~ & ~  & ~ & ~ & ~ & ~ & ~  & ~ & ~ & ~ & ~ & ~ & ~ \\
    
~~~No Failure  &  N/A    & N/A & N/A   & \textbf{97.84} & 88.23 & \textbf{91.94} & 97.81 & \textbf{88.53} & 91.43 & 97.53 & 87.75 & 88.44 & 96.92 & 84.60 & 85.79 \\
~~~\high & 96.32 & \textbf{86.78} & \textbf{89.54} & 96.64 & 85.03 & 89.50 & \textbf{97.07} & 85.87 & 88.70 & 97.04 & 86.66 & 86.28 & 96.52 & 84.01 & 84.13 \\
~~~\med  &  95.81 & 81.61 & \textbf{86.16} & 94.96 & 80.30 & 85.81 & 95.70 & 81.92 & 84.59 & \textbf{95.86} & \textbf{84.92} & 82.97 & 95.38 & 82.99 & 81.55\\
~~~\low     & \textbf{91.95} & 77.46 & 78.60 & 89.36 & 70.32 & \textbf{78.82} & 90.58 & 73.16 & 77.03 & 91.06 & \textbf{79.93} & 76.97 & 90.67 & 79.35 & 76.65\\
    \bottomrule
    \end{tabular}
    \label{tab:failout-results}
\end{table*}

{\bf 2. \em What is the optimal rate of failout?}  In this ablation experiment, we investigate the effect of failout by setting the rate of failout to fixed rates of 5\%, 10\%, 30\%, 50\%, and a varying rate of ``Failure,'' where the failout rate for a physical node is equal to its probability of failure during inference. \Cref{tab:failout-results} illustrates the impact of failout rate in \schemeNameResilinet{}. {\bf\em Key result 3}: ResNet (R) seems to favor {\em Failure} failout rate, and MobileNet (M) favors higher failout rates of around 30\%.. {\bf\em Key observation 3}: we hypothesize that, since a significant portion of the DNN is dropped during training when using failout, higher failout rate results in lower accuracy, as opposed to standard dropout. 
%In standard dropout, the dropout rates are generally higher than 5\%, because individual neurons are dropped with such rates, which are a small part of the DNN. 
%In \high{} reliability setting, the variable rate of Reliability has the highest performance gain. This suggests that the optimal failout rate in \schemeNameResilinet{} may depend on the reliability setting of the network, or even on the neural network architecture. 
{\bf\em Key observation 4}: based on our preliminary experiments, we conclude that the optimal failout rate should be seen as a hyperparameter, and be tuned for the experiment.

\begin{figure*}[t]
\centering
    \begin{minipage}{0.44\textwidth}
    \centering
    \includegraphics[width=\linewidth]{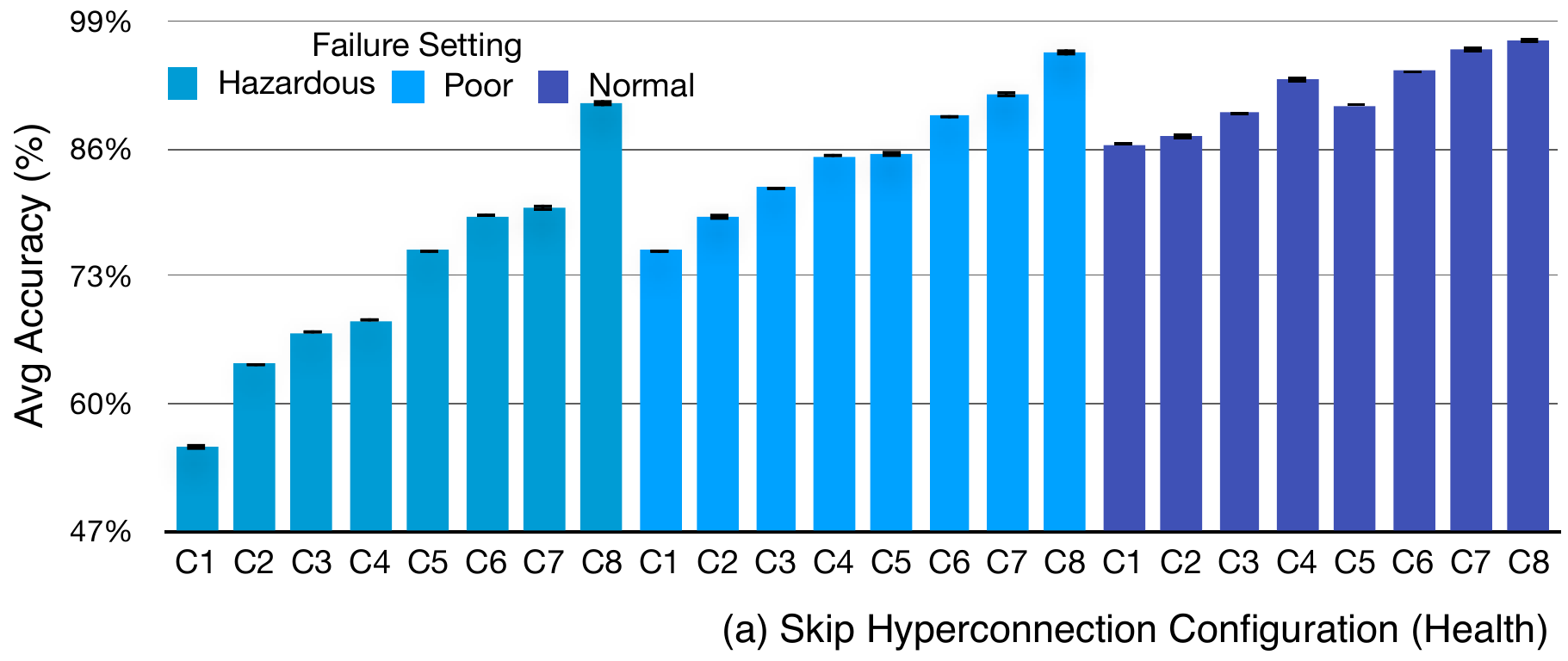}
    \includegraphics[width=\linewidth]{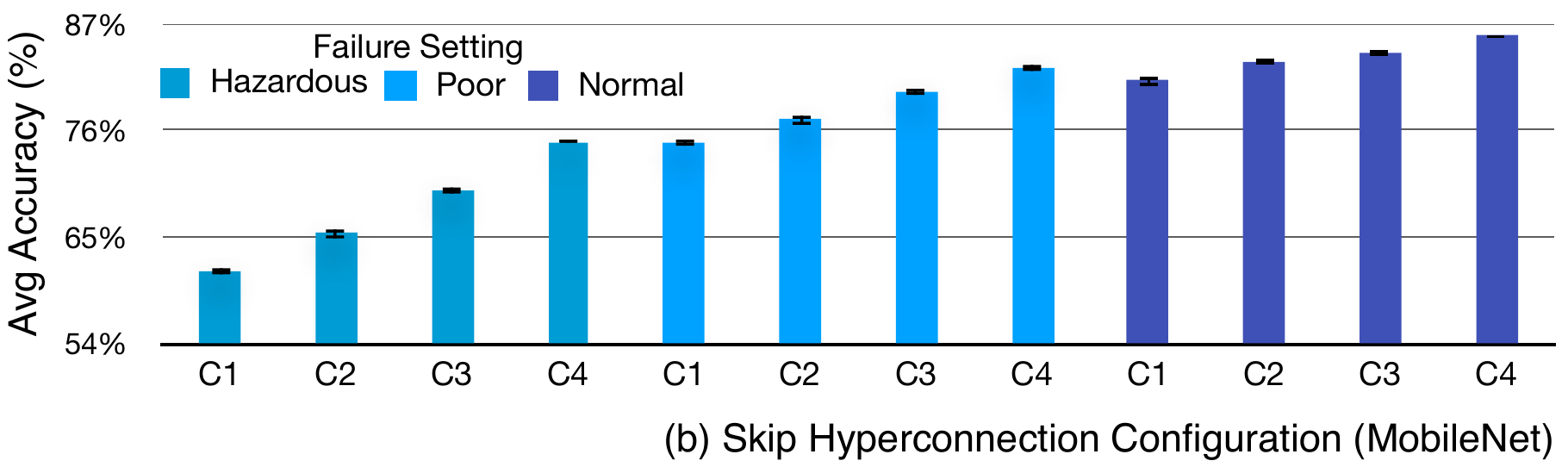}
    \includegraphics[width=\linewidth]{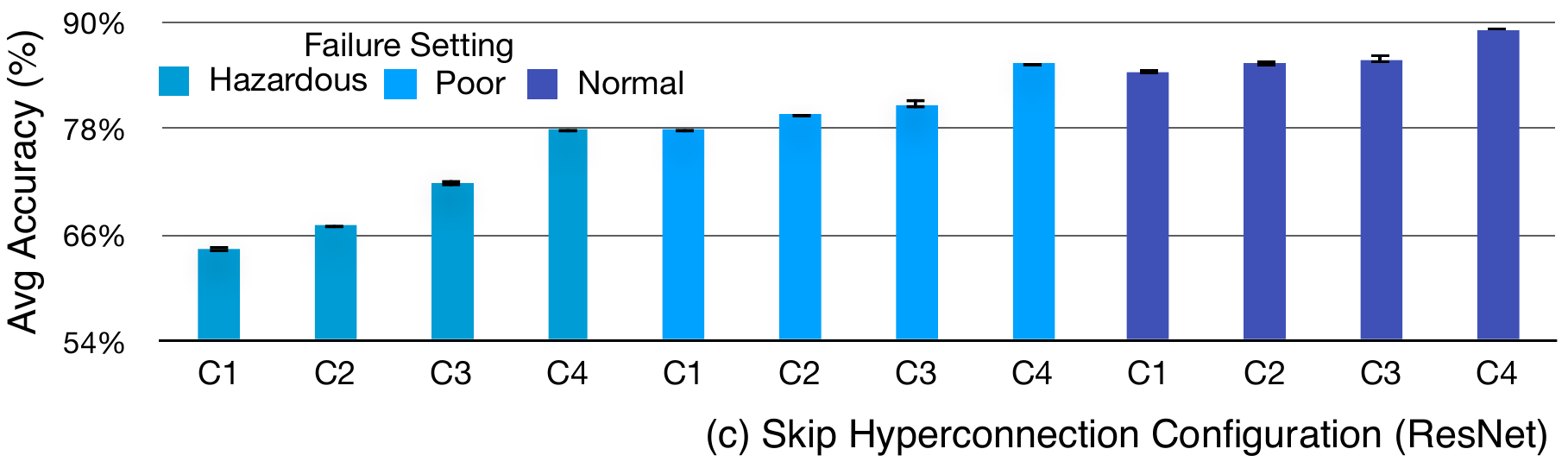}
    \end{minipage}
    \begin{minipage}{0.15\textwidth}
    \centering
\tiny
 \vspace{-0.2in}
 \setlength{\tabcolsep}{2pt}
\begin{tabular}{@{}lr@{}}
\midrule
Config & Skip Hyperconnection\\
C1 & None \\
C2 & $n$\scalebox{\subscriptscale}{$_2$} \\
C3 & $n$\scalebox{\subscriptscale}{$_1$} \\
C4 & $i$ \\
C5 & $n$\scalebox{\subscriptscale}{$_1$}, $n$\scalebox{\subscriptscale}{$_2$} \\
C6 & $n$\scalebox{\subscriptscale}{$_2$}, $i$ \\
C7 & $n$\scalebox{\subscriptscale}{$_1$}, $i$ \\
C8 & All \\ \midrule 
 ~ & ~ \\[0.22in] \midrule
% Cfg & Present Skip HC\\ 
% C1 & $z$\scalebox{\subscriptscale}{$_4$} \\
% C2 & $z$\scalebox{\subscriptscale}{$_3$}, $z$\scalebox{\subscriptscale}{$_4$} \\
% C3 & $z$\scalebox{\subscriptscale}{$_2$}, $z$\scalebox{\subscriptscale}{$_4$} \\
% C4 & $z$\scalebox{\subscriptscale}{$_4$}, $z$\scalebox{\subscriptscale}{$_7$} \\
% C5 & $z$\scalebox{\subscriptscale}{$_2$}, $z$\scalebox{\subscriptscale}{$_7$} \\
% C6 & $z$\scalebox{\subscriptscale}{$_1$}, $z$\scalebox{\subscriptscale}{$_5$}, $z$\scalebox{\subscriptscale}{$_6$}, $z$\scalebox{\subscriptscale}{$_7$} \\
% C7 & $z$\scalebox{\subscriptscale}{$_1$}, $z$\scalebox{\subscriptscale}{$_2$}, $z$\scalebox{\subscriptscale}{$_5$}, $z$\scalebox{\subscriptscale}{$_6$},  $z$\scalebox{\subscriptscale}{$_7$}\\
% C8 & All \\ \midrule
%  ~ & ~ \\[0.1in] \midrule
Config & Skip Hyperconnection\\ 
C1 & None \\
C2 & $m$\scalebox{\subscriptscale}{$_1$}\\
C3 & $i$\\
C4 & All \\ \midrule
 ~ & ~ \\[0.22in] \midrule
Config & Skip Hyperconnection\\ 
C1 & None \\
C2 & $i$\\
C3 & $z$\scalebox{\subscriptscale}{$_1$}\\
C4 & All \\ \midrule

\end{tabular}
    \end{minipage}
    \caption{Ablation studies for analyzing sensitivity of \schemeNameResilinet{}'s skip hyperconnections in (a) \health{} experiment, (b) MobileNet experiment, (c) ResNet experiment. The charts show average top-1 accuracy (with error bars showing standard deviation). The tables to the right of the charts show the present skip hyperconnections in each skip hyperconnection configuration. (Notation: letters indicate the source physical node of the corresponding skip hyperconnection)}
    \label{fig:sensitivity-results}
\end{figure*}

{\bf 3. \em Which skip hyperconnections are more important?} 
It is important to see which skip hyperconnections in \schemeNameResilinet{} are more important, thereby contributing more to the resiliency of the distributed neural network. This is helpful for certain scenarios in which having all skip hyperconnections is not possible (e.g. due to the cost of establishing new connections, or some communication constraints). To perform these experiments, we shut down (i.e. disconnect) a certain configuration of skip hyperconnections while keeping other skip hyperconnections active and every experiment setting the same, to see changes in the performance. The results are presented in \cref{fig:sensitivity-results}. The bars show the average top-1 accuracy of 10 runs, under different ``configs'' in which a certain combination of skip hyperconnections are shut down. The present skip hyperconnections are shown in the tables next to the bar charts. Letters in the tables indicate the source physical node of the skip hyperconnection. In the \health{} experiment, since there are three skip hyperconnections in the distributed neural network, there are eight possible configurations of skip hyperconnections (``Config 1'' through ``Config 8''). Similarly, in the experiments with MobileNet and ResNet-18, we consider all four configurations, as we have two skip hyperconnections.

% \begin{table*}[h]
% \centering
% \scriptsize
% \caption{Present skip hyperconnections in each skip hyperconnection configuration. (Notation: letters indicate the source node of the corresponding skip hyperconnection)} 
% % \vspace{-4}
% \begin{tabular}{@{}lcccccccc@{}}
% \toprule
% Configuration & 1 & 2 & 3 & 4 & 5 & 6 & 7 & 8 \\ \midrule 
% Health      & None & $f_2$ & $e_1$ & $e_1, f_2$ & $g_1$ & $g_1, f_2$ & $g_1, e_1$ & All \\
% Camera   & $e_1, e_2, e_3, e_4$ & $e_1, f_2$ & $e_1, f_3$ & $f_2, f_3$ & $f_2$ & $e_1, e_2, e_3, e_4, f_2$ & $e_2, e_3, e_4, f_2, f_3$ & All \\ 
% \bottomrule
% \end{tabular}
% \label{tab:config-key}
% \end{table*}

In the \health{} (\cref{fig:sensitivity-results}a), we can see a uniform accuracy gain, when going from Config 1 towards Config 8. We can also see that, by looking at Config 2 through Config 4, if only one skip hyperconnection is allowed in a scenario, it should be the skip hyperconnection from input to $n_2$ (labeled as $i$). This is also evident when comparing Config 5 and Config 6: the skip hyperconnection from input to $n_2$ is more important. In the \low{} reliability scenario, a proper subset of two skip hyperconnections can achieve up to a 24\% increase in average accuracy (Config 1 vs. Config 6). {\bf\em Key result 4}: this hints that individual skip hyperconnections are more important when there are more failures in the network. In the experiment with MobileNet, we also observe a uniform accuracy increase, when going from Config 1 towards Config 4. We can see that the skip hyperconnection from input to $m_2$ is more important that the skip hyperconnection from $m_1$ to $m_3$ (Config 2 vs. Config 3). Nonetheless, if both skip hyperconnections are present (Config 4), the performance is at its peak. Comparably, in the experiment with ResNet (\cref{fig:sensitivity-results}c), we can see that the skip hyperconnection from node $z_1$ to $z_3$ is more important than the skip hyperconnection from input to $z_2$. We can also see that, when we have all the skip hyperconnections, the performance of the distributed DNNs are at their peak.

% In \camera{} (\cref{fig:sensitivity-results}b), when the reliability setting is high (e.g. \high{}), it is hard to tell which skip hyperconnections are more important. It is also interesting to see that, in \high{} reliability setting, when only the skip hyperconnection in Config 1 is present, accuracy is high, and addition of some skip hyperconnections can even reduce the accuracy (e.g. see Config 1 vs. Config 2 and 3). Same trend is also evident in \med{} reliability settings. However, in \low{} reliability setting, we can see the addition of the skip hyperconnection from $z_7$ to $z_9$ increases the accuracy (see Config 4 through Config 8, where $z_7$ is present). Another increase can be seen when skip hyperconnections labeled by their source nodes $z_5$ and $z_6$ are added (see Config 6 through Config 8).

This ablation study demonstrates that, by searching for a particular \textit{important} subset of skip hyperconnections in a distributed neural network, especially in the \low{} reliability scenarios, we can achieve a large increase in the average accuracy. We point the interested reader to \cite{further-ablation-studies, residual2, residual3} for more in-depth analyses of representations and functions induced by skip connections in neural networks.

%This concludes the discussion of our experiments. In the next section we explain the state of the art in this direction, and we position our work's novelty in the literature.

\section{Related Work} \label{related}
%The related work in this space can be categorized in the following groups.

{\bf a. Distributed Neural Networks}. Federated Learning is a paradigm that allows clients collaboratively train a shared global model \cite{fed1, fed2, fed3}. Similarly, distributed training of neural networks has received significant attention \cite{tensorFlow, Pytorch, microsoft}. Resilient distributed training against adversaries is studied in ~\cite{pmlr-v80-chen18l, Damaskinos}. Nevertheless, inference in distributed neural networks is less explored, although application scenarios that need ongoing and long inference tasks are emerging \cite{harvard, morshed2017deep, EdgeEye, hotedge-distributed, ChuangHu, dey1}. 

{\bf b. Neural Network Fault Tolerance}. A related concept to failure is {\em fault}, which is when units or weights become defective (i.e. stuck at a certain value, or random bit flip). Studies on fault tolerance of neural networks date back to the early 90s, and are limited to mathematical models of small neural networks (e.g. neural networks with one hidden layer or unit-only and weight-only faults) \cite{mehrotra1994fault, Bolt92faulttolerance, phatak1995complete}. 
%  However, none of these works consider failure of physical nodes that potentially cause the failure of a large group of neural network units and weights.

{\bf c. Neural Network Robustness}. A line of research related to our study is robust neural networks \cite{goodfellow-robust, Szegedy-robust, cisse2017parseval, bastani2016measuring, el2017robustness}. Robustness in neural networks has gained considerable attention lately, and is especially important when the neural networks are to be developed in commercial products. These studies are primarily focused on adversarial examples, examples that are only slightly different from correctly classified examples drawn from the data distribution. Despite the relation to our study, we are not focusing on the robustness of neural networks to adversarial examples. We study resiliency of distributed DNN inference in the presence of failure of a large group of neural network units. \schemeNameDFG{} framework in \cite{previous} uses skip hyperconnections for failure-resiliency of distributed DNN inference. We showed how \schemeNameResilinet{} differs from \schemeNameDFG{} in skip hyperconnections setup, and in its novel use of failout to provide greater failure-resiliency.

{\bf d. Regularization Methods}. Some regularization methods that implicitly increase robustness are dropout \cite{normal-dropout}, dropConnect \cite{dropconnect}, DropBlock \cite{Dropblock}, zoneout \cite{Zoneout}, cutout \cite{cutout}, swapout \cite{Swapout}, and stochastic depth \cite{layerwise}. Although there are similarities between failout and these methods in terms of the regularization procedure, these methods largely differ in spirit from ours. In particular, although during training, dropout turns off neurons and dropConnect discards weights, they both enable an ensemble of models for regularization. On the other hand, failout shuts down an entire physical node in a distributed neural network to simulate actual failures in the physical network, for providing failure-resiliency. Stochastic depth is a procedure to train very deep neural networks effectively and efficiently. The focus of zoneout, DropBlock, swapout, and cutout is on regularizing recurrent neural networks and CNNs, while they are not designed for failure-resiliency. 

\section{Conclusion} \label{Conclusion}
Federated Learning utilize deep learning models for training or inference without accessing raw data from clients. Similarly, we presented \schemeNameResilinet{}, a framework for providing failure-resiliency of distributed DNN inference that combines two concepts: skip hyperconnections and failout. We saw how \schemeNameResilinet{} can improve the failure-resiliency of distributed MLPs and distributed CNNs. We also observed experimentally that, the weight of hyperconnections may not change the performance of distributed DNNs if the hyperconnections weights are chosen in certain range. We also observed that the rate of failout should be seen as a hyperparameter and be tuned. Finally, we observed that some skip hyperconnections are more important than others, especially under more extreme failure scenarios.

 {\bf Future Work}: We view \schemeNameResilinet{} as an important first step in studying failure-resiliency in distributed DNNs. This study opens several paths for related research opportunities. Firstly, it is interesting to study the distributed DNNs that are both horizontally and vertically distributed. Moreover, finding optimal hyperconnection weights through training (not through heuristics) may be a future research direction. Finally, instead of having only skip hyperconnection to bypass a node, we can have a {\em skip layer}, a layer to approximate the neural components of a failed physical node. 

% \section*{References}

\section{Broader and Ethical Impact}

{\bf Energy and Resources}: \schemeNameResilinet{} may take longer to converge, due to its failout regularization procedure. Moreover, if a distributed DNN is already trained, it needs to be re-trained with skip hyperconnections and failout; though, the training can be done offline. Additionally, some hyper-parameter tuning may be needed during training. These training settings depend on the availability of large computational resources that necessitate similarly substantial energy consumption \cite{energy}. We did not prioritize computationally efficient hardware and algorithms in the experiment. Nevertheless, if \schemeNameResilinet{} is deployed and is powered by renewable energy and, the impacts of the hyperparameter tuning will be offset over a long period of time. Regarding bandwidth usage, \schemeNameResilinet+ also increases the use of bandwidth due the activity of the skip hyperconnections both during training and inference.  

{\bf Bias}: Secondly, as the large scale deployment of powerful deep learning algorithms becomes easier and more practical, the number of new applications that will use the infrastructure will undoubtedly grow. With the new applications, there is a risk that models are over-fit and biased to a particular setting. The bias and over-fit may impact people (e.g. when the model may not be ``fair''), especially when more people become users of such applications. Although we do not provide solutions or countermeasures to these issues, we acknowledge that this type of research can implicitly carry a negative impact in the future regarding the issues described above. Follow-up work focusing on applications must therefore include this type of consideration.

\bibliographystyle{ACM-Reference-Format}
\bibliography{Master}

\clearpage
\section{Supplementary Material}

\subsection{Different configurations of hyperconnections}
In this paper, all of the experiments are conducted on vertically distributed DNNs, as they are more common form of distributed DNNs. Nevertheless, one could imagine a distributed DNN that is both vertically and horizontally distributed. For example, when a DNN is used for image-based defect detection in a factory or automatic recognition of parts during product assembly, maybe it is distributed vertically and horizontally for dispersed presence \cite{harvard, previous}. In these cases, the horizontal distribution of DNN helps to extend the DNN to multiple regions which may are geographically distributed. As an example, in a case where inference runs across geographically distributed sites, the first few layers of the distributed DNN can be duplicated (horizontally) and placed on the corresponding physical nodes in those sites, so that they can perform the forward pass on the first few layers. One (or more) downstream node can then combine the immediate activations sent from those physical nodes and send the combined activation to the upper layers of the DNN.

\begin{figure}[h]
\centering
     \begin{minipage}{.05\linewidth}
    ~
    \end{minipage}
     \begin{minipage}{.1\linewidth}
    \centering
    \vspace{0.1in}
    \subfloat[b][]{\includegraphics[width=\linewidth]{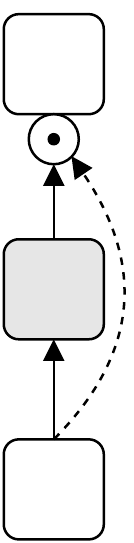}
        \label{fig:config1}}
    \end{minipage}
    \begin{minipage}{.05\linewidth}
    ~
    \end{minipage}
    \begin{minipage}{.22\linewidth}
    \centering
    \subfloat[][]{\includegraphics[width=\linewidth]{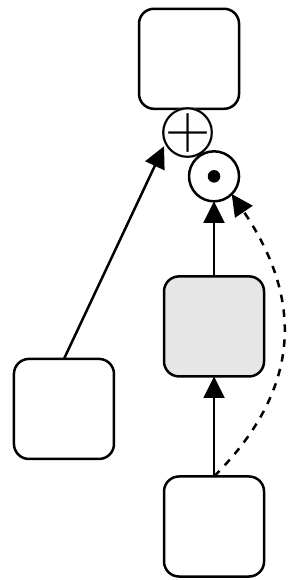}
        \label{fig:config2}}
    \end{minipage}
    \begin{minipage}{.05\linewidth}
    ~
    \end{minipage}
    \begin{minipage}{.2\linewidth}
    \centering
    \vspace{0.03in}
    \subfloat[][]{\includegraphics[width=\linewidth]{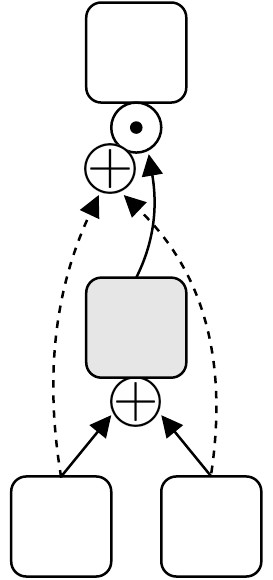}
        \label{fig:config3}}
    \end{minipage}
     \caption{\schemeNameResilinet{}'s configurations of hyperconnections. Boxes denote physical nodes and arrows denote hyperconnections. The shaded physical node is the ``{\em failing node}'' that undergoes failure (during inference) or failout (during training). (a) The {\em failing node} is the only child of its parent and has only one child, (b) The {\em failing node} is not the only child of its parent and has one child, (c) The {\em failing node} is the only child of its parent and has more than one child.}
	\label{fig:resilinet-config}
\end{figure}

\Cref{fig:resilinet-config} shows \schemeNameResilinet{}'s different configurations of hyperconnections. \Cref{fig:config1} shows a vertically distributed DNN, and \cref{fig:config2} and \cref{fig:config3} show a distributed DNN that is both vertically and horizontally distributed. Other distributed DNN architectures could be constructed based on the combination of these three basic hyperconnection configurations. In \schemeNameResilinet{}, skip hyperconnections are active only during failure or failout; Thus, in \cref{fig:resilinet-config}, the symbol $\odot$ represents this behavior, which was defined previously in the paper as follows: in $X_{i+1}=Y_i\odot Y_{i-1}$, when node $v_i$ (shaded in gray in \cref{fig:config1}) is alive $X_{i+1}=Y_i$, and when node $v_i$ fails, $X_{i+1}=Y_{i-1}$. The symbol $\oplus$ simply denotes addition. For instance, in \cref{fig:config2}, the input to the top node is the sum of the output of the node on the left, and the output of one of the nodes on the right, depending on if the gray node is alive or not. Recall that in \schemeNameResilinet{}+, we replace the symbol $\odot$ with the symbol $\oplus$. Thus, in the figure for hyperconnection configurations of \schemeNameResilinet{}+, we would just have a single symbol $\oplus$ on the input to the top node that adds all the incoming outputs.

\subsection{Different Structure of distributed DNN} 

In this subsection, to verify our claims regarding the superior performance of \schemeNameResilinet{}, we consider different partitions of DNNs onto distributed physical nodes and measure their performance. For this ablation study, we consider the distributed MLP in \health{} experiment, and the distributed MobileNet. For the MLP in \health{} experiment, instead of the 1$\rightarrow$1$\rightarrow$2$\rightarrow$3$\rightarrow$4 partition that we considered in the paper, we experiment with partition 1$\rightarrow$2$\rightarrow$3$\rightarrow$2$\rightarrow$3. For MobileNet, instead of the 1$\rightarrow$3$\rightarrow$5$\rightarrow$5 partition, we experiment with partition 2$\rightarrow$2$\rightarrow$4$\rightarrow$6. 

\begin{figure}[t]
    \centering
    \centering
    \includegraphics[width=1\linewidth]{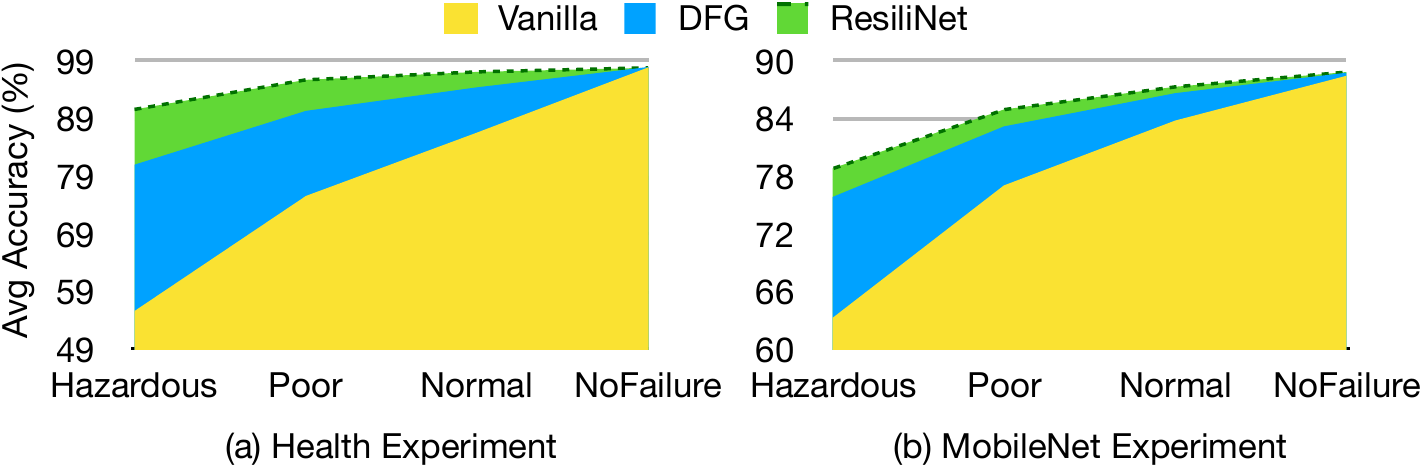}
    \caption{Average performance under new partition}
    \label{fig:different-partition-results}
\end{figure}

The results of our experiments with these new two partitions are depicted in \cref{fig:different-partition-results}. We can see that, \schemeNameResilinet{} consistently outperforms both \schemeNameDFG{} and vanilla, and this verify our claims regarding the superior performance of \schemeNameResilinet{} in a new distributed DNN partition. We also experimented with other partitions, and observed the same trends. 

%\subsection{Accuracy in ResNet Experiments}
%This subsection explores the accuracy of 

\clearpage
\section{Frequently Asked Questions}

In this section, we hope to answer the frequently asked questions by the reviewers of this paper and the audience of this work when we present it.

\subsection{What exactly the failure settings are and why nodes fail with those
particular probabilities?}

{\bf Answer:} The failure settings are representations of real-world failure conditions, simply to show that under more severe conditions, \schemeNameResilinet{}'s benefit is more evident. And regarding failure values, the nodes closer to the cloud have less failure rate. That is why in \cref{tab:surv-configs} the failure rate of nodes closer to the cloud are less than those farther from the cloud. See the next question.

\subsection{What is a typical failure probability in a real system?}

{\bf Answer:} For cloud, it is almost 0\% (99.999\% availability). Devices closer to the cloud and data centers, e.g., backbone nodes, have also relatively high availability; according to Meza, Justin, et al. ``{\em A large scale study of data center network reliability}'', published in Proceedings of the Internet Measurement Conference, 90\% of such nodes have a mean time between failures (MTBF) of 3521 hours and a mean time to repair (MTTR) of 71 hours. Thus, the availability of those nodes is around 98\%. Devices closer to end-user are expected to have less availability, of around 92\%-98\%. These numbers are in accordance with the failure setting ``\high'' in the paper. Similarly, other failure settings have lower reliability.

\subsection{How does time to detect a failure impact \schemeNameResilinet{} in a real deployment?}

{\bf Answer:} Note that with the heartbeat mechanism (mentioned in the paper), time to detect a failure is in order of seconds (at most minutes), while the time to fix a node is in order of hours; per Meza, Justin, et al. ``{\em A large scale study of data center network reliability}'', published in Proceedings of the Internet Measurement Conference, 90\% of backbone nodes have a mean time to repair of 71 hours. 

\subsection{What is the failure model of physical nodes?}

{\bf Answer:} Our failure model is crash-only non-Byzantine.

\subsection{Do skip hyperconnections increase bandwidth usage?}

{\bf Answer:} In \schemeNameResilinet{}, skip hyperconnections are active only during failure (and during failout), and they only route the blocked information flow. Note that skip hyperconnection's advantage is in routing information during failure, that is otherwise not possible. The bandwidth for the routed information over the failed node is the same as when there is no failure (skip hyperconnection only finds a detour). 

\subsection{What is the bandwidth savings of \schemeNameResilinet{}?}

{\bf Answer:} Here we measure the bandwidth savings of \schemeNameResilinet{} compared to \schemeNameDFG{}. In \schemeNameDFG{}, skip hyperconnections are always active during training and inference. However, in \schemeNameResilinet{}, skip hyperconnections are only activated when a node fails. To obtain the bandwidth savings, we measure the size of the data that is passed among nodes at any given time. \Cref{tab:bandwidth} shows the bandwidth savings in different experiments.

\begin{table}[t]
\scriptsize
% \vspace{-0.1in}
\caption{Bandwidth savings of \schemeNameResilinet{}}
\begin{tabular}{@{}lccc@{}}
\toprule
Experiment    & Dist. MLP & Dist. MobileNet & Dist. ResNet-18\\ \midrule
Dataset       & UCI Health & ImageNet~~~~ CIFAR-10 & CIFAR-10\\ \midrule
Bandwidth savings & $40.35\%$ & $47.06\%$~~~~ $46.23\%$ & $40.61\%$\\ 
\bottomrule
\end{tabular}
\label{tab:bandwidth}
\end{table}

\subsection{How about experiments on actual sensors and devices?}

{\bf Answer:} The training procedure in \schemeNameResilinet{} can be performed completely off-line (not during runtime), and later the learned network can be deployed. Hence, the actual deployment would not impact the training. It is also worth noting that, with real implementation, since the failure rates and time to repair during inference are most larger, the benefits of \schemeNameResilinet{}
would be more evident. Recall that the benefit of \schemeNameResilinet{} is more evident in more extreme failure conditions.

\subsection{Is this distributed inference framework related to Federated Learning?}

{\bf Answer:} This is a similar setting to Federated Learning, but it differs from it significantly. Federated Learning aims to train distributed deep models without sharing the raw data with the centralized server. In Federated Learning, model parameters are available to all participants and data are locally distributed among devices. Conversely, in distributed inference of neural networks, model parameters are split by different nodes. In other words, by partitioning the network and distributing it across several physical nodes, activations and gradients are exchanged between physical nodes, rather than raw data.

\subsection{Is there any assumption on the distribution of data on different nodes?}

{\bf Answer:} No, we do not assume any distribution of data (e.g. on IoT nodes).

\subsection{What is the suggested/optimal range for failout rate?}

{\bf Answer:} Based on \cref{tab:failout-results} the rate of failout should be chosen less than 0.5.

\subsection{Why is failout related to regularization?}

{\bf Answer:} The main similarities can be drawn to dropout regularization, which serves as the main inspiration to this work. In dropout, by dropping units at random in layers of neural network, the learned weights are regularized to show a more robust behavior that does not depend entirely on output of one particular set of neurons, but on all of them as a whole. Similarly, in failout we randomly drop a physical node, which allows regularizing weights in the following nodes, making them less dependent on this particular node (which can fail).

\subsection{How are skip hyperconnections activated? What is the protocol to reconfigure?}

{\bf Answer:} \schemeNameResilinet{} simply routes information through the skip hyperconnection, when a failure is detected. Once failed node is fixed, \schemeNameResilinet{} turns off the skip hyperconnection and sends the data normally. 

\subsection{Why not permit a systems approach of dynamically provisioning a replica to replace the faulty node?}

{\bf Answer:} We think this approach is not as efficient as simply starting a new connection upon failure of a node. In the systems approach, we would have some sort of a backup node or a reserved resource, to start upon failure; in \schemeNameResilinet{} we just open a connection (e.g., TCP connection) after failure detection for starting a skip hyperconnection.

\subsection{How much do skip hyperconnections add to the size of the model and training time?}

{\bf Answer:} When hyperconnections do not normally introduce new parameters their size is negligible. Only when hyperconnections need to match the dimensions (e.g., in CNNs), they introduce new parameters. The training time is in fact slightly reduced, as training time is usually faster with networks with residual connection (skip hyperconnections are similar to residual connections). 

\end{document}